%% file: main.tex
\documentclass[11pt,a4paper]{article}
\usepackage[hyperref]{acl2021}
\usepackage{times}
\usepackage{latexsym}

\usepackage{microtype}

\usepackage[english]{babel} 
\usepackage[utf8]{inputenc}
\usepackage[T5]{fontenc}
\usepackage{amsmath}            
\usepackage{epstopdf}           
\usepackage{flushend}           
\usepackage{amssymb}
\usepackage[figuresright]{rotating}
\usepackage{subfigure}
\usepackage{xcolor}
\usepackage{times}
\usepackage{latexsym}
\usepackage{float}
\usepackage{multirow}
\usepackage{makecell}
\usepackage{microtype}
\usepackage{hyperref}
\usepackage{xcolor, soul}
\sethlcolor{green}

\aclfinalcopy

\title{VLSP 2021 - ViMRC Challenge: Vietnamese Machine Reading Comprehension}

\author{Kiet Van Nguyen$^{1,2,4}$, Son Quoc Tran$^{3,5}$, Luan Thanh Nguyen$^{1,2,4}$, Tin Van Huynh$^{1,2,4}$, \\\bf Son T. Luu$^{1,2,4}$, Ngan Luu-Thuy Nguyen$^{1,2,4}$ \\
  $^{1}$University of Information Technology, Ho Chi Minh City, Vietnam \\
  $^{2}$Vietnam National University, Ho Chi Minh City, Vietnam \\
  $^{3}$Denison University, USA \\
  \texttt{$^{4}$\{kietnv,luannt,tinhv,sonlt,ngannlt\}@uit.edu.vn,$^{5}$tran\_s2@denison.edu}}

\date{}

\begin{document}
\maketitle

\begin{abstract}
One of the emerging research trends in natural language understanding is machine reading comprehension (MRC) which is the task to find answers to human questions based on textual data. Existing Vietnamese datasets for MRC research concentrate solely on answerable questions. However, in reality, questions can be unanswerable for which the correct answer is not stated in the given textual data. To address the weakness, we provide the research community with a benchmark dataset named UIT-ViQuAD 2.0 for evaluating the MRC task and question answering systems for the Vietnamese language. We use UIT-ViQuAD 2.0 as a benchmark dataset for the challenge on Vietnamese MRC at the Eighth Workshop on Vietnamese Language and Speech Processing (VLSP 2021). This task attracted 77 participant teams from 34 universities and other organizations. In this article, we present details of the organization of the challenge, an overview of the methods employed by challenge participants, and the results. The highest performances are 77.24\% in F1-score and 67.43\% in Exact Match on the private test set. The Vietnamese MRC systems proposed by the top 3 teams use XLM-RoBERTa, a powerful pre-trained language model based on the transformer architecture. The UIT-ViQuAD 2.0 dataset motivates researchers to further explore the Vietnamese machine reading comprehension task and related tasks such as question answering, question generation, and natural language inference.
\end{abstract}

\input{section/1-introduction.tex}
\input{section/2-related_works.tex}
\input{section/3-task.tex}
\input{section/4-dataset.tex}

\input{section/5-results.tex}
\input{section/6-resultanalysis}
\input{section/7-conclusion.tex}

\section*{Acknowledgments}

The authors would like to thank the team of aihub.vn\footnote{\url{https://aihub.vn/}}, and the annotators for their hard work to support the VLSP 2021 - ViMRC Challenge. The VLSP Workshop was supported by organizations: VINIF, Aimsoft, Zalo, Bee, and INT2, and universities: VNU-HCM University of Information Technology, VNU University of Science, and VNU University of Engineering and Technology. Kiet Van Nguyen was funded by Vingroup JSC and supported by the Master, PhD Scholarship Programme of Vingroup Innovation Foundation (VINIF), Institute of Big Data, code VINIF.2021.TS.026.

\bibliographystyle{acl_natbib}
\bibliography{reference}



\end{document}

%% file: section/1-introduction.tex
\section{Introduction}

Machine Reading Comprehension (MRC) is an emerging and challenging task of natural language understanding that computers can read and understand texts and then find correct answers to any questions. Recently, many MRC shared tasks \cite{cui2018span,fisch2019mrqa, zheng2021semeval} and benchmark corpora \cite{richardson-etal-2013-mctest,rajpurkar-etal-2016-squad,rajpurkar-etal-2018-know,joshi2017triviaqa,trischler-etal-2017-newsqa,kocisky-etal-2018-narrativeqa,lai-etal-2017-race,reddy-etal-2019-coqa} have attracted a range of researchers from academia and industry. Therefore, significant progress has been exploited in building computational models for semantics based on deep neural networks and transformers over the last ten years \cite{seo2016bidirectional,devlin2019bert,conneau2019unsupervised,van2021vireader}. The datasets and models are studied in resource-rich languages such as English and Chinese. Until now, no MRC shared task or challenge has been organized for Vietnamese, which motivates us to organize a challenge for Vietnamese machine reading comprehension.

\begin{figure*}
    \centering
    \includegraphics[width=1\linewidth]{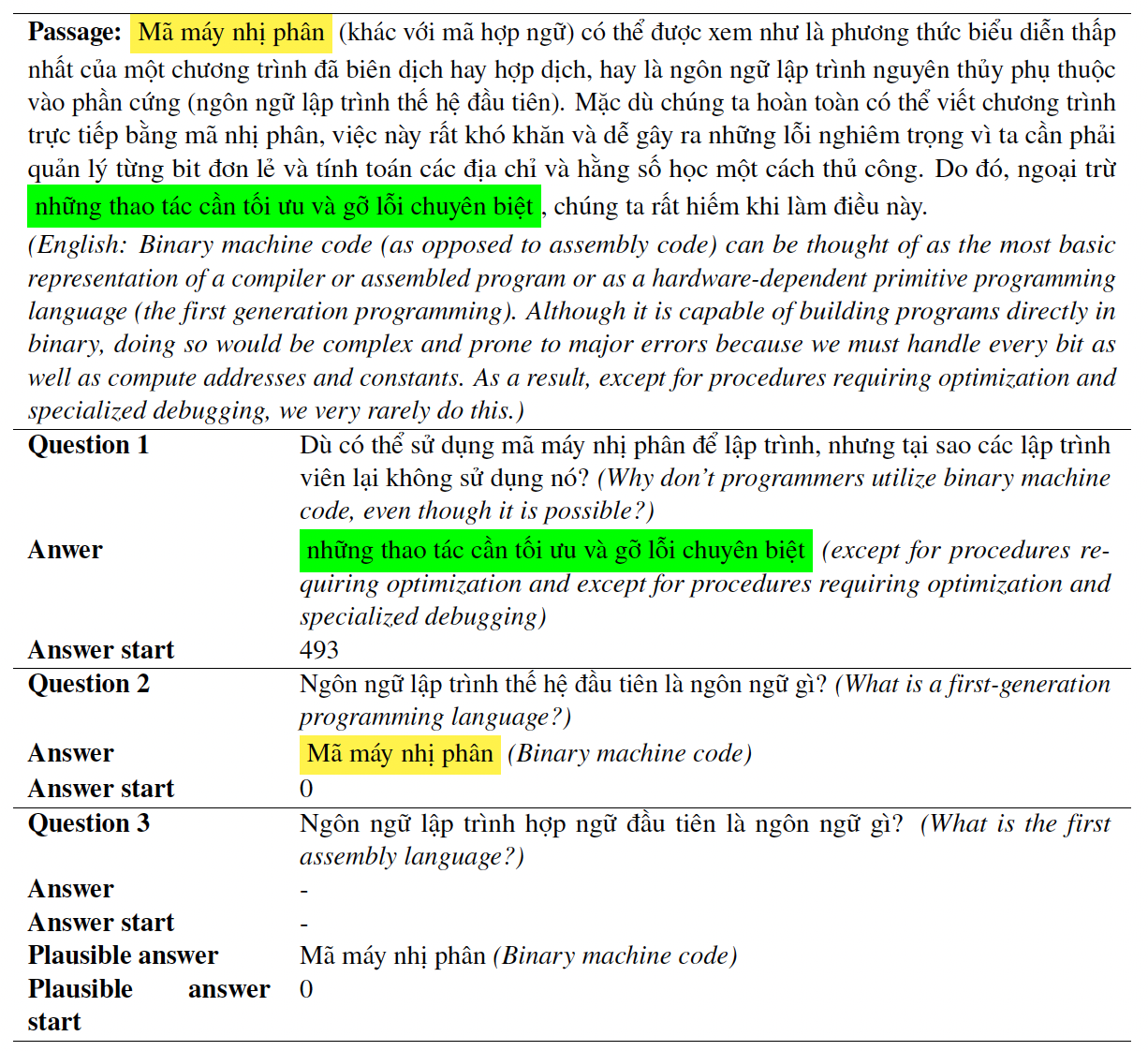}
    \caption{Several passage-question-answer triples extracted from the dataset.}
    \label{fig:example}
\end{figure*}

We introduce the VLSP 2021 - ViMRC Challenge: Vietnamese Machine Reading Comprehension. We hope to use this challenge to examine the capabilities of state-of-the-art deep learning and transformer models to represent and simulate machine reading comprehension for Vietnamese texts. Inspired by machine reading comprehension benchmarking \cite{rajpurkar-etal-2018-know}, we design this challenge of Vietnamese reading comprehension, in which computers are given a document D as well as a human question Q$_i$ to comprehend. In this work, we construct UIT-ViQuAD 2.0, a new dataset that combines answerable questions from the previous version of UIT-ViQuAD (UIT-ViQuAD 1.0 \cite{nguyen-etal-2020-vietnamese}) with over 12K unanswerable questions for the same passages. Figure \ref{fig:example} illustrates two such examples.

The participating teams made 590 total submissions within the official VLSP-2021 evaluation period. We introduce the challenge and present a summary for the evaluation in this paper.

In this paper, we have three main contributions described as follows.

\begin{itemize}
    \item Firstly, we constructed UIT-ViQuAD 2.0, a Vietnamese dataset for the span-extraction reading comprehension task which contains nearly 36,000 human-annotated questions including unanswerable and answerable. Unanswerable questions are added to increase the linguistic diversity in machine reading comprehension and question answering.
    \item Secondly, we organize the VLSP 2021 - ViMRC Challenge for evaluating MRC and question answering models in Vietnamese at the VLSP 2021. Our baseline system obtains 63.03\% and 60.34\% in F1-score on the public and private test sets, respectively, and there is no model of participating teams that pass 78\% (in F1-score) on the private test set, which indicates our dataset is challenging and requires the development of MRC models for the Vietnamese.
    \item UIT-ViQuAD 2.0 could also be a good resource for multilingual and cross-lingual research purposes when studied along with other MRC and QA datasets.
    
\end{itemize}

The following is how the rest of the article is organized. In Section \ref{relatedwork}, we provide a brief overview of the background and relevant studies. We introduce the VLSP 2021 - ViMRC Challenge in Section \ref{vlspmrcsharedtask}. Our new dataset (UIT-ViQuAD 2.0) is presented in detail in Section \ref{datasetconstruction}. Section \ref{experiment} presents the systems and results proposed by participating teams. In Section \ref{resultanalysis}, we provide further analysis of the challenge results. Finally, Section \ref{conclusion} summarizes the findings of the VLSP 2021 - ViMRC Challenge and suggests future research directions.

%% file: section/2-related_works.tex
\section{Background and Related Work}
\label{relatedwork}

\begin{table*}[!h]
\centering
\begin{tabular}{l l r c c}
\hline
\bf Dataset        & \bf Language   & \bf Size  & \bf Answerable & \bf Unaswerable \\ \hline
SQuAD1.1 \cite{rajpurkar-etal-2016-squad}       & English    & 100k+ & \checkmark            &             \\ 
SQuAD2.0 \cite{rajpurkar-etal-2018-know}       & English    & 150k+  & \checkmark            & \checkmark             \\ \hline
KorQuAD \cite{lim2019korquad1}        & Korean     & 70k+  & \checkmark            &             \\ \hline
SberQuAD \cite{braslavski2020sberquad}       & Russian    & 50k+  & \checkmark            &             \\ \hline
CMRC-2018 \cite{cui2018span}       & Chinese    & 20k+  & \checkmark            &             \\ \hline
FQuAD1.1 \cite{d2020fquad} & French     & 60k+  & \checkmark            &             \\ 
FQuAD2.0 \cite{heinrich2021fquad2}       & French     & 60k+  & \checkmark            & \checkmark              \\ \hline
UIT-ViNewsQA \cite{van2020new}       & Vietnamese & 23k+  & \checkmark            &             \\ 
UIT-ViQuAD 1.0 \cite{nguyen-etal-2020-vietnamese} & Vietnamese & 22k+  & \checkmark            &             \\ 
UIT-ViQuAD 2.0 (Ours) & Vietnamese & 35k+  & \checkmark            & \checkmark             \\ \hline
\end{tabular}
\caption{Benchmark of existing reading comprehension datasets, including UIT-ViQuAD.}\label{lbl:datasets}
\end{table*}

Machine Reading Comprehension (MRC) has attracted many researchers in developing machine learning-based MRC models after the introduction of SQuAD (a large-scale and high-quality dataset) \cite{rajpurkar-etal-2016-squad}. The growth in human-annotated datasets and computing capabilities are key factors behind the dramatic progress in the machine reading comprehension models. Particularly, many of datasets are constructed for evaluating the machine reading comprehension task including extraction-based MRC datasets (SQuAD \cite{rajpurkar-etal-2016-squad}, SQuAD 2.0 \cite{rajpurkar-etal-2018-know}, TriviaQA \cite{joshi2017triviaqa}, and NewsQA \cite{trischler-etal-2017-newsqa}), abstractive MRC dataset (NarrativeQA \cite{kocisky-etal-2018-narrativeqa}, RECAM \cite{zheng-etal-2021-semeval}), multiple-choices datasets (RACE \cite{lai-etal-2017-race} and MCTest \cite{richardson-etal-2013-mctest}), and conversational reading comprehension dataset (CoQA \cite{reddy-etal-2019-coqa} and ViCoQA \cite{9352127}). In addition to the creation of the MRC datasets, various neural network techniques \cite{seo2016bidirectional,devlin2018bert, conneau2019unsupervised,van2021deep} have been presented and made significant progress in this field. Table \ref{lbl:datasets} shows the comparison of different MRC datasets.

Various efforts to create Vietnamese MRC datasets have been conducted. UIT-ViQuAD \cite{nguyen-etal-2020-vietnamese}, UIT-ViNewsQA \cite{van2020new} are two corpora for the extraction-based machine reading comprehension task in Vietnamese language. Besides, two Vietnamese QA systems \cite{xlmrserini,bertqas} were developed with automatic reading comprehension techniques. In addition, ViMMRC \cite{9247161} and ViCoQA \cite{10.1007/978-3-030-88113-9_44} are two Vietnamese corpora for multiple-choices reading comprehension and conversational reading comprehension, respectively. Besides, a few MRC and QA methods have been studied on Vietnamese MRC datasets, such as BERT \cite{9352127}, ViReader \cite{van2021vireader}, XLMRQA \cite{xlmrserini}, and ViQAS \cite{bertqas}.

Ultimately, SQuAD 2.0 \cite{rajpurkar-etal-2018-know}, and NewsQA \cite{trischler-etal-2017-newsqa} are two corpora claiming the challenge of unanswerable questions in machine reading comprehension tasks, which are similar to our shared task. In general, extraction-based MRC requires computer understanding and retrieving the correct answer from the reading texts, which can evaluate the comprehension of the computer's natural language texts. However, the computer not only answers given questions as usual but also knows which questions are unanswerable. Our purpose in the shared task is to construct a dataset to evaluate the ability of the computer on both answerable and unanswerable questions for the extraction-based machine reading comprehension task.  

%% file: section/3-task.tex
\section{The VLSP 2021 - ViMRC Challenge}
\label{vlspmrcsharedtask}
\subsection{Task Definition}

This task aims to enable the ability of computers to understand natural language texts and answer relevant questions from users. The task is defined as below:
\begin{itemize}
    \item \textbf{Input:} Given a text T = \{t$_1$, ..., t$_n$\}and a question Q = \{q$_1$, ..., q$_m$\} which can be answerable or unanswerable.
    \item \textbf{Output:} An answer A = [a$_s$, a$_e$], where $0 \leq a_s  \leq a_e \leq n$ can be a span ${t_{a_s}, ..., t_{a_e}}$ extracted directly from T or empty if no answer is found. 
\end{itemize}


The answers returned by the system are represented as answer spans by character level and are extracted from the reading text. The spans begin with an index indicating the location of the answer in the reading text. The end of the spans is an index determined by the sum of the start index and the length of the answers text. Nevertheless, the question in this task consists of answerable and unanswerable questions (as described in Figure \ref{fig:example}), which is more difficult than the ViQuAD dataset \cite{nguyen-etal-2020-vietnamese}. 


According to Figure \ref{fig:example}, the first and the second questions are answerable questions. The answers are directly extracted from the reading passage (highlighted by colors in the reading passage. The blue one is the answer for the first question, and the red one is the answer for the second question). The third question is unanswerable, however, according to \citet{rajpurkar-etal-2018-know}, the plausible answers are added to the dataset to make it more diverse and create the challenge for current machine reading comprehension to enhance the ability of computers for understanding natural languages.

\subsection{Evaluation Metrics}

Following the evaluation metrics on SQuAD2.0 \cite{rajpurkar-etal-2018-know}, we use EM and F1-score as evaluation metrics for Vietnamese machine reading comprehension. These evaluation metrics are described as below:
\begin{itemize}
    \item Exact Match (EM): If the characters of the MRC system's predicted answer exactly match the characters of (one of) the gold standard answer(s), EM = 1 for each question-answer pair; otherwise, EM = 0. The EM metric is a strict all-or-nothing measurement, with a score of 0 for a single character error. If the method predicts any textual span as an answer when evaluating against an unanswerable question, the question receives a zero score.
    
    \item F1-score: F1-score is a popular metric for natural language processing and is also used in machine reading comprehension. F1-score estimated over the individual tokens in the predicted answer against those in the gold standard answers. The F1-score is based on the number of matched tokens between the predicted and gold standard answers.
\end{itemize}

The final ranking is evaluated on the private test set, according to the F1-score (EM as a secondary metric when there is a tie).

\subsection{Schedule and Overview Summary}

Table \ref{importantdate} shows important dates of the VLSP 2021 - ViMRC Challenge. It lasted for two months, during which the participating teams spent 27 days developing the models.

\begin{table}[H]
\centering

\begin{tabular}{l l}
\hline
\multicolumn{1}{c}{\textbf{Time}} & \multicolumn{1}{c}{\textbf{Phase}} \\ \hline
October 1st                         & Trial Data                          \\ 
October 5th                         & Public test                         \\ 
October 25th                        & Private test                        \\ 
October 27th                        & Competition end                     \\ \hline
November 15th                       & Submission deadline                 \\ 
December 15th                       & Notification of acceptance          \\ 
December 28th                       & Camera-ready due                    \\ \hline
\end{tabular}
\caption{Schedule of the VLSP 2021 - ViMRC Challenge.}
\label{importantdate}
\end{table}

Besides, Table \ref{tbl_participate_sumary} describes an overview of participants who joined the competition. To get access to the system, each team must nominate a delegate, and register with the organizers. Only delegates of teams can submit the result to the system (as shown on the leaderboard). 

\begin{table}[H]
    \centering
    
    \begin{tabular}{l r}
        \hline
        \textbf{Metric} & \textbf{Value} \\
        \hline
        \#Registration Teams & 77 \\
        \#Joined Teams & 42 \\
        \#Signed Data Agreements & 42\\
        \#Submitted Teams & 24 \\
        \#Paper Submissions &  6\\
        \hline
    \end{tabular}
    \caption{Participation summary of the VLSP 2021 - ViMRC Challenge.}
    \label{tbl_participate_sumary}
\end{table}

\begin{table}[H]
    \centering
     
    \begin{tabular}{l p{1.2cm} p{1.2cm} p{1.2cm}}
        \hline
        & \textbf{Public Test} & \textbf{Private Test} & \textbf{Overall} \\
        \hline
        Total Entries & 551 & 39 & 590 \\
        Highest F1 & 84.24 & 77.24 & 84.24 \\
        Highest EM & 77.99 & 67.43 & 77.99 \\
        Mean F1 & 70.70 & 60.96 & 66.37 \\
        Mean EM & 61.13 & 50.47 & 56.39 \\
        Std. F1 & 12.34 & 23.38 & 18.52 \\
        Std. EM & 12.57 & 20.82 & 17.38 \\
        \hline
    \end{tabular}
    \caption{Results overview of the VLSP 2021 - ViMRC Challenge.}
    \label{tbl_results_sumary}
\end{table}

Finally, Table \ref{tbl_results_sumary} shows the statistical information about the results of participants by F1 and EM scores. Overall, the highest EM score is not higher than 78 percent, while the highest F1 score is 84.24 percent. Both the highest F1 and EM scores come from the public test. However, the results on the private test set are lower. Notably, the standard deviation of results by F1 and EM scores on the private test set is significantly higher than the public test set, which means the results between participating teams are different.  

%% file: section/4-dataset.tex
\section{Dataset Construction}
\label{datasetconstruction}
We proposed a new dataset named UIT-ViQuAD 2.0 for this task, the latest version of the Vietnamese Question Answering Dataset. This dataset includes questions from the first version of UIT-ViQuAD \cite{nguyen-etal-2020-vietnamese} and nearly 13,000 newly human-generated questions which are unanswerable (see Section \ref{unanswerable}) and answerable (see Section \ref{answerable}).

Instead of generating unanswerable questions from scratch like SQuAD 2.0 \cite{rajpurkar-etal-2018-know}, we transform answerable questions into unanswerable questions. We randomly sample one-half of answerable questions in the original dataset and ask our annotators to transform these questions into unanswerable ones, which are impossible to answer given the information of the passage. The answers for answerable questions are then used as the plausible answers for unanswerable questions. This ensures that the unanswerable questions are similar to answerable ones, and the quality of plausible answers for unanswerable questions is high enough for further research into the behavior of Question Answering models.

\subsection{Generating Unanswerable Questions}
\label{unanswerable}
To generate unanswerable questions, we do a strict process of two phases: (1) unanswerable question creation and (2) unanswerable question validation.
\subsubsection{Unanswerable Question Creation}
We hire 13 high-quality annotators for the process of generating unanswerable questions, most of whom have experience in annotating different datasets in Vietnamese Natural Language Processing. Our hired annotators are carefully trained in 6 phases in 10 days with 30 questions each phase. In the first 2 phases, we mainly focus on getting our annotators familiar with the task. In the next 4 phases, annotators are asked to create questions with a diverse range of unanswerable categories. We did this by having our 13 annotators transform the same set of questions. Then, when more than two annotators have the same way of transforming an answerable question into an unanswerable one, these annotators will be asked to transform that question again. The result of this process is that there are many categories of unanswerable questions in our dataset, such as Antonym, Overstatement, Understatement, Entity Swap, Normal Word Swap, Adverbial Clause Swap, 
Modifiers Swap. This proposes new challenges to Vietnamese Machine Reading Comprehension researchers. Table \ref{tab:categories} presents categories of unanswerable questions in UIT-ViQuAD 2.0.

We include all answerable questions, besides newly generated unanswerable ones, from the previous version of our dataset. This gives us a dataset with the proportion of roughly one unanswerable question per 2 answerable questions. Table \ref{overview} summarizes the dataset's overall statistics.

\begin{table*}[p]

\resizebox{\textwidth}{!}{
\begin{tabular}{p{3cm}ll}

\hline
\textbf{Reasoning}    & \textbf{Description}                                                                                                   & \multicolumn{1}{c}{\textbf{Example}}                                                                                      \\ \hline
Antonym               & Antonym used                                                                                                           & \begin{tabular}{p{14cm}}\textbf{Sentence}: Vào năm 1171, Richard khởi hành đến Aquitaine với mẹ mình và Henry phong ông là Công tước xứ Aquitaine theo yêu cầu của Eleanor.  (\textit{In 1171, Richard departed for Aquitaine with his mother, and Henry made him Duke of Aquitaine at Eleanor's request})\\ 
\textbf{Original question}: Richard khởi hành đến Aquitaine với mẹ vào năm nào?  (\textit{In what year did Richard depart for Aquitaine with his mother?}) \\ Unanswerable question: Richard khởi hành từ Aquitaine với mẹ vào năm nào?  (\textit{In what year did Richard depart from Aquitaine with his mother?}) \end{tabular}                                                                                                                                                                                                                                                                                                                                                                                                                                                                                                                                    
\\ \hline
Overstatement         & \begin{tabular}{p{5cm}} Word that has similar meaning but with a higher shades of meaning is used\end{tabular}   & \begin{tabular}{p{14cm}}\textbf{Sentence}: Ngày 9 tháng 11 năm 1989, vài đoạn của Bức tường Berlin bị phá vỡ, lần đầu tiên hàng ngàn người Đông Đức vượt qua chạy vào Tây Berlin và Tây Đức. (\textit{On November 9, 1989, several parts of the Berlin Wall were collapsed, and for the first time thousands of East Germans crossed into West Berlin and West Germany.)}\\  \textbf{Original question}: Bức tường Berlin đã bị sụp đổ một vài đoạn vào ngày nào? (\textit{On which date were some parts of Berlin Wall collapsed?})\\ \textbf{Unanswerable question}: Bức tường Berlin đã bị sụp đổhoàn toàn vào ngày nào? (\textit{On which date was Berlin Wall completely collapsed?})\end{tabular}                                                                                                                                                                                                                                                                                                                                                                                                                                                                                   \\ \hline
Understatement        & \begin{tabular}{p{5cm}}Word that has similar meaning but with a lower shades of meaning is used\end{tabular}     & \begin{tabular}{p{14cm}}\textbf{Sentence}: Quân đội Nhật Bản chiếm đóng Quảng Châu từ năm 1938 đến 1945 trong chiến tranh thế giới thứ hai. (\textit{The Japanese army captured Guangzhou from 1938 to 1945 during the second world war.})\\  \textbf{Original question}: Khi Chiến tranh Thế giới thứ hai xảy ra thì Quảng Châu bị nước nào chiếm đóng? (\textit{During the World War II, Guanzong was captured by which country?})\\ \textbf{Unanswerable question}: Khi Chiến tranh Thế giới thứ hai xảy ra thì Quảng Châu bị nước nào đe dọa? (\textit{During the World War II, Guanzong was attacked by which country?})\end{tabular}                                                                                                                                                                                                                                                                                                                                                                                                                                                                                                                                                \\ \hline
Entity Swap           & Entity replaced by other entity                                                                                        & \begin{tabular}{p{14cm}}\textbf{Sentence}: Là cảng Trung Quốc duy nhất có thể tiếp cận được với hầu hết các thương nhân nước ngoài, thành phố này đã rơi vào tay người Anh trong chiến tranh nha phiến lần thứ nhất. (\textit{As the only Chinese port accessible to most foreign merchants, the city fell to the British during the First Opium War.})\\  \textbf{Original question}: Trong cuộc chiến nào thì Anh Quốc đã chiếm được Quảng Châu? (\textit{In which war did Britain capture Guangzhou?})\\ \textbf{Unanswerable quetion}: Trong cuộc chiến nào thì Nhật đã chiếm được Quảng Châu? (\textit{In which war did Japan capture Guangzhou?})\end{tabular}                                                                                                                                                                                                                                                                                                                                                                                                                                                                                                                     \\ \hline
Normal Word Swap      & \begin{tabular}{p{5cm}}A normal word replaced by another normal word\end{tabular}                                & \begin{tabular}{p{14cm}}\textbf{Sentence}: Sự phát hiện của Hofmeister năm 1851 về các thay đổi xảy ra trong túi phôi của thực vật có hoa {[}...{]} (\textit{Hofmeister's discovery in 1851 of changes occurring in the embryo sac of flowering plants {[}...{]}})\\ \\ \textbf{Original question}: Năm 1851 nhà sinh học Hofmeister đã tìm ra điều gì ở thực vật có hoa? (\textit{In 1851, the biologist Hofmeister discovered what in flowering plants?})\\ \textbf{Unanswerable question}: Năm 1851 nhà sinh học Hofmeister đã công nhận điều gì ở thực vật có hoa? (\textit{In 1851, the biologist Hofmeister accepted what in flowering plants?})\end{tabular}                                                                                                                                                                                                                                                                                                                                                                                                                                                                                                                                    \\ \hline
Adverbial Clause Swap & \begin{tabular}{p{5cm}}Adverbial clause replaced by another adverbial clause related to the context\end{tabular} & \begin{tabular}{p{14cm}}\textbf{Sentence}: Trước đó Phạm Văn Đồng từng giữ chức vụ Thủ tướng Chính phủ Việt Nam Dân chủ Cộng hòa từ năm 1955 đến năm 1976. Ông là vị Thủ tướng Việt Nam tại vị lâu nhất (1955–1987). Ông là học trò, cộng sự của Chủ tịch Hồ Chí Minh. (\textit{Pham Van Dong previously held the position of Prime Minister of the Democratic Republic of Vietnam from 1955 to 1976. He was the longest-serving Prime Minister of Vietnam (1955-1987). He was a student and collaborator of President Ho Chi Minh.})\\ \textbf{Original question}: Giai đoạn năm 1955-1976, Phạm Văn Đồng nắm giữ chức vụ gì? (\textit{In the period 1955-1976, what position did Pham Van Dong hold?}) \\ \textbf{Unanswerable question}: Khi là cộng sự của chủ tịch Hồ Chí Minh, Phạm Văn Đồng nắm giữ chức vụ gì? (\textit{As a collaborator of President Ho Chi Minh, what position did Pham Van Dong hold?})\end{tabular}                                                                                                                                                                                                                                                     \\ \hline
Modifiers Swap      & \begin{tabular}{p{5cm}} Modifier of one word in the given context is used for another word\end{tabular}        & \begin{tabular}{p{14cm}}\textbf{Sentence}: Các phần mềm giáo dục đầu tiên trong lĩnh vực giáo dục đại học (cao đẳng) và tập trung được thiết kế chạy trên  máy tính đơn (hoặc các thiết bị cầm tay). Lịch sử của các phần mềm này được tóm tắt trong SCORM 2004 2nd edition Overview (phần 1.3) (\textit{The first educational software in the field of higher education (college) and concentration was designed to run on a single computer (or portable devices). The history of these software is summarized in SCORM 2004 2nd edition Overview (section 1.3).}) \\  \textbf{Original question}: Lịch sử của các phần mềm giáo dục đầu tiên trong lĩnh vực giáo dục đại học (cao đẳng) được tóm tắt, ghi nhận ở đâu? (\textit{Where did the history of the first educational software in the field of higher education} (\textit{college}) \textit{was summarized and recorded?}) \\ \textbf{Unanswerable quetion}: Lịch sử của các phần mềm giáo dục trong lĩnh vực giáo dục đại học (cao đẳng) được tóm tắt, ghi nhận đầu tiên ở đâu? (\textit{Where did the history of the educational software in the field of higher education (college) was first summarized and recorded?})\end{tabular} \\ \hline

\end{tabular}
}
\caption{Categories of unanswerable questions in UIT-ViQuAD 2.0.}
\label{tab:categories}
\end{table*}

\subsubsection{Unanswerable Question Validation}
Before publishing the dataset for the evaluation campaign, we have carefully validated newly unanswerable questions following the procedure inspired by \citet{nguyen-etal-2020-vietnamese}. To help annotators gradually be better at generating new unanswerable questions, after generating every 3,000 unanswerable questions, we asked our annotators to self-validate the questions that they have generated before and write short documents to reflect on their errors. This effort minimizes the possibility that our annotators repeat their errors too many times.

To further reduce the error rate in our unanswerable questions, we have a separate phase of cross-validating after finishing creating 12,000 unanswerable questions. We hired ten annotators who had generated over 1,000 unanswerable questions during the phase of generating new samples for this phase. This effort helped filter out the annotators who have little experience in annotating unanswerable questions to reduce the noise during the validation phase. Our team then investigated and confirmed every error detected by annotators. To maximize the probability of detecting errors in newly generated unanswerable questions, we provide our annotators with incentives to carefully check for the errors in the dataset as we additionally reward them on each error they correctly detect. 

\subsection{Additional Difficult Answerable Questions}
\label{answerable}

In addition to answerable questions from UIT-ViQuAD 1.0, we also hire five annotators, who have experiences in doing researches with Vietnamese natural language processing and clearly understand different reasoning skills \cite{sugawara-etal-2017-evaluation} that is important to evaluate the comprehension ability of models to annotate more challenging answerable questions, which requires models more reasoning ability to correctly answer. The selected annotators are then encouraged to spend at least 3 minutes per question. When generating this set of questions, our purpose is to propose more challenges to researchers in the VLSP 2021 Evaluation Campaign and encourage further analysis on the effects of unanswerable questions in future works.

\subsection{Overview Statistics of UIT-ViQuAD 2.0}

\begin{table*}[ht]
\centering

\resizebox{\textwidth}{!}{
\begin{tabular}{ l r r r r }
\hline
       & \textbf{Train} & \textbf{Public Test} & \textbf{Private Test} & \textbf{All}\\
\hline
 \#Articles & 138 & 19 & 19 & 176 \\  
 \#Passages & 4,101 & 557 & 515 & 5,173\\
 \#Total questions & 28,457 & 3,821 & 3,712 & 35,990 \\
 \#Unanswerable questions & 9,217 & 1,168 & 1,116 & 11,501 \\
 Average passage length & 179.0 & 167.6 & 177.3 & 177.6 \\
 Average answerable question length & 14.6 & 14.3 & 14.7 & 14.6\\
 Average unanswerable question length & 14.7 & 14.0 & 14.5 & 14.6 \\
 \hline
\end{tabular}
}
\caption{Overview Statistics of UIT-ViQuAD 2.0.}
\label{overview}
\end{table*}

The general statistics of the datasets are
given in Table \ref{overview}. UIT-ViQuAD 2.0 comprises 35,990 question-answer-passage triples (including 9,217 unanswerable . The organizers provide training, public test, and private test sets for the participating teams. For public and private test sets, we only provide passages and their questions without answers to the teams.

%% file: section/5-results.tex
\section{Systems and Results}
\label{experiment}

\subsection{Baseline System}
Following \citet{devlin2019bert}, we adopt transfer learning based on BERT (Bidirectional Encoder Representations from Transformers) for our baseline system. To adapt our dataset, we slightly modify the run squad.py script\footnote{\url{https://github.com/google-research/bert/blob/master/run_squad.py}} while keeping the majority of the original code. mBERT is trained in 104 languages, including Vietnamese. In addition, we use the transformers library by Hugging Face\footnote{\url{https://huggingface.co/}} to fine-tune mBERT for our question-answering dataset. We fine-tuned the parameters to suit our dataset in the training process as well as the model evaluation process.

For the baseline system, we used an initial learning\_rate of 3e-5 with a batch\_size of 32 and trained for two epochs. The max\_seq\_length and doc\_stride are set to 384 and 128.

\subsection{Challenge Submissions}

The AIHUB platform\footnote{\url{https://aihub.vn)}} was used to manage all submissions. We received entries from 24 teams for the public test, while for the private test, we received submissions from 19 teams. The systems using the pre-trained language model XLM-R achieve SOTA results. Six of these teams had their system description papers submitted. Each of them is briefly described below.

\subsubsection{The vc-tus team}

With addressing unanswerable questions, \citet{vlspmrc1} presents a novel Vietnamese reading comprehension system based on Retrospective Reader \cite{zhang2021retrospective}. Furthermore, they concentrate on improving answer extraction ability by utilizing attention mechanisms efficiently and boosting representation capacity through semantic information processing. They also offer an ensemble strategy for achieving significant improvements in single model results. Their method won the first place in the VLSP 2021 -- ViMRC Challenge.

\subsubsection{The ebisu\_uit team}
\citet{vlspmrc4} presents a novel method for training Vietnamese reading comprehension. To tackle the Machine reading comprehension task in Vietnamese, they apply BLANC (BLock AttentioN for Context prediction) \cite{seonwoo2020context} on pre-trained language models. With this strategy, this model produced good results. This approach achieved 77.22 percent of F1-score on the private test with the MRC task at the VLSP 2021 -- ViMRC Challenge, placing the second rank overall.

\subsubsection{The F-NLP team}
To learn the correlation between a start answer index and an end answer index in pure-MRC output prediction, \citet{vlspmrc3} present two types of joint models for answerability prediction and pure-MRC prediction with/without a dependence mechanism. They also use ensemble models and a verification approach that involves choosing the best answer from among the top K answers offered by different models.

\subsubsection{The UIT-MegaPikachu team}

\citet{vlspmrc5} propose a new system which employs simple yet highly effective method. The system uses a strong pre-trained language model (PrLM) XLM-RoBERTa \cite{conneau2019unsupervised}, combined with filtering results from multiple outputs to produce the final result. This system generated around 5-7 outputs and chose the answer with the highest number of repetitions as the final predicted answer. 

\subsubsection{The UITSunWind team}

\citet{vlspmrc2} present the description of a new approach to solve this task at the VLSP 2021 -- ViMRC Challenge. A novel system MRC4MRC using XLM-RoBERTa includes two main components. On the public-test set, the MRC4MRC based on the XLM-RoBERTa pre-trained language model achieves 79.13 percent in F1-score and 69.72 percent in Exact Match. Despite being among the top 5 models, the EM-based performance on answerable questions is the best on the private test. The XLM-RoBERTa language model outperforms the strong PhoBERT language model in their experiments.

\subsubsection{The HN-BERT team}

\citet{vlspmrc6} offer an unsupervised passage selector that shortens a given passage while retaining answers in related passages. In the corpus of the VLSP 2021 -- ViMRC Challenge, they also applied a variety of experimental techniques, such as unanswerable question sample selection and different adversarial training methodologies, which enhanced performance by 2.5 percent in EM and 1 percent in F1-score.

\begin{table*}[h]
\centering

\begin{tabular}{lllllrr}
\hline
\multicolumn{3}{c}{\textbf{Public Test Phase}} &
  \textbf{} &
  \multicolumn{3}{c}{\textbf{Private Test Phase}} \\ \hline
\multicolumn{1}{l|}{\textbf{}} &
  \textbf{F1} &
  \textbf{EM} &
   &
  \multicolumn{1}{l|}{\textbf{}} &
  \multicolumn{1}{l}{\textbf{F1}} &
  \multicolumn{1}{l}{\textbf{EM}} \\ \hline
\multicolumn{1}{l|}{\textbf{Human}}      & 87.335          & 81.818          &  & \multicolumn{1}{l|}{\textbf{Human}}    & 82.849          & 75.500          \\ \hline
\multicolumn{1}{l|}{\textbf{NLP\_HUST}}  & \textbf{84.236} & 77.728          &  & \multicolumn{1}{l|}{\textbf{vc-tus}}   & \textbf{\colorbox{yellow}{77.241}} & 66.137          \\
\multicolumn{1}{l|}{\textbf{NTQ}} &
  \textbf{84.089} &
  \textbf{77.990} &
   &
  \multicolumn{1}{l|}{\textbf{ebisu\_uit}} &
  \textbf{77.222} &
  \textbf{\colorbox{yellow}{67.430}} \\
\multicolumn{1}{l|}{\textbf{ebisu\_uit}} & \textbf{82.622} & 73.698          &  & \multicolumn{1}{l|}{\textbf{F-NLP}}    & \textbf{76.456} & 64.655          \\
\multicolumn{1}{l|}{vc-tus}              & 81.013          & 71.316          &  & \multicolumn{1}{l|}{UIT-MegaPikachu}   & 76.386          & 65.329          \\
\multicolumn{1}{l|}{F-NLP}               & 80.578          & 70.662          &  & \multicolumn{1}{l|}{SDSOM}             & 75.981          & 63.012          \\
\multicolumn{1}{l|}{SDSOM}               & 79.594          & 69.092          &  & \multicolumn{1}{l|}{UITSunWind}        & 75.587          & 64.871          \\
\multicolumn{1}{l|}{UITSunWind}          & 79.130          & 69.720          &  & \multicolumn{1}{l|}{Big Heroes}        & 74.241          & 61.126          \\
\multicolumn{1}{l|}{UIT-MegaPikachu}     & 78.637          & 68.804          &  & \multicolumn{1}{l|}{914-clover}        & 73.027          & 61.853          \\
\multicolumn{1}{l|}{914-Clover}          & 78.515          & 69.013          &  & \multicolumn{1}{l|}{NTQ}               & 72.863          & 60.938          \\
\multicolumn{1}{l|}{Big Heroes}          & 78.491          & 68.150          &  & \multicolumn{1}{l|}{Hey VinMart}       & 70.352          & 57.786          \\
\multicolumn{1}{l|}{PhoKho-UIT}          & 75.894          & 65.533          &  & \multicolumn{1}{l|}{PhoKho-UIT}        & 70.198          & 58.378          \\
\multicolumn{1}{l|}{HN-BERT}             & 75.842          & 63.544          &  & \multicolumn{1}{l|}{HN-BERT}           & 70.100          & 56.466          \\
\multicolumn{1}{l|}{Hey VinMart}         & 75.759          & 64.590          &  & \multicolumn{1}{l|}{Deep-NLP}          & 69.220          & 59.429          \\
\multicolumn{1}{l|}{Deep-NLP}            & 74.767          & 66.789          &  & \multicolumn{1}{l|}{ABC}               & 63.625          & 55.280          \\
\multicolumn{1}{l|}{ABC}                 & 69.287          & 57.864          &  & \multicolumn{1}{l|}{\textbf{BASELINE}} & \textbf{60.338} & \textbf{49.353} \\
\multicolumn{1}{l|}{ct-nlp}              & 68.971          & 58.859          &  & \multicolumn{1}{l|}{}                  &                 &                 \\
\multicolumn{1}{l|}{tpp}                 & 68.484          & 57.786          &  & \multicolumn{1}{l|}{}                  &                 &                 \\
\multicolumn{1}{l|}{S-NLP}               & 67.589          & 65.140          &  & \multicolumn{1}{l|}{}                  &                 &                 \\
\multicolumn{1}{l|}{\textbf{BASELINE}}   & \textbf{63.031} & \textbf{53.546} &  & \multicolumn{1}{l|}{}                  &                 &                 \\ \hline
\end{tabular}
\caption{Final results on the public and private test sets. Participating teams are ranked by their highest F1-score.}
\label{tab:result}
\end{table*}

\subsection{Human Performance}
To estimate the human performance of this task, we employ a team to answer a set of 100 samples from the public test set and 100 samples from the private test set. There are four annotators, and two of them work on each set doing the same thing.

In each instance, we have a passage with a question. The annotator must answer the question using the information in the passage. If there is no answer, it means the question is unanswerable, and then mark "true" in the field "is unanswerable".
Following the answering phase, we compute the human accuracy by F1-score and exact match scores for both public and private tests.

To calculate human performance, we use the method given in SQuAD2.0 \cite{rajpurkar-etal-2018-know}. We have four responses per question in the ground truth. Thus we choose the final ground truth by majority voting and prefer the shortest answer to be the last ground truth, as explained in SQuAD2.0. After obtaining the gold response, we compute the F1 and EM scores in pairs of human-answering and gold answers with the two annotators who previously answered on the public test set. Then, by averaging the results of the two annotators, we compute the final F1 and EM scores of human performance on the public test. The computation is carried out on the private test in the same manner. As a result, the final F1 and EM scores of human performance are 87.34\% and 82.85\% on the public test set, respectively, and 81.82\% and 75.50\% on the private test set.

\subsection{Experimental Results}

According to our statistics, a total of 24 teams registered to submit their results. These teams from prestigious universities, companies, and organizations participate in the Vietnamese Machine Reading Comprehension task of the VLSP 2021 - ViMRC Challenge. And then, out of the 24 teams participating in the development phase of their system on the public test we selected 18 teams that excelled against the baseline to further evaluate their system in the private test.

The results of the teams in the two rounds are aggregated and shown in Table \ref{tab:result}. The ranking results of the team are based on F1 points for both rounds. In the public test round, our mBERT baseline model achieved 63.03\% on F1 and 53.55\% on EM. There were 18 teams with results that outperformed the results of the baseline according to F1. Overall, 14 teams with F1 scores above 70\% and 5 teams with over 80\%. Specifically, we found the top three teams in the public test, NLP\_HUST, NTQ, and ebisu\_uit, with F1 results of 84.24\%, 84.09\%, and 82.63\%, respectively. It can be seen that the results of the two top teams in the rankings have very close results. The difference between these two teams is not more than 0.2\%. Additionally, the NTQ team's model scored slightly lower in F1 than NLP\_HUST, but their model achieved the highest EM performance of 77.99\%.

Regarding the private test round, the baseline model's results achieved an F1 score of 60.34\% and 49.35\% on the EM score. Out of the 18 teams that passed the public test, 14 continued to participate in evaluating their system on the private test set. There have been many unexpected changes in the results of the teams' submissions, especially the way the top three teams appeared. While only placing in 5th with 81.01\% of F1 on the public test set, team vc-tus took 1st position in the private test round with an F1 score of 77.24\%. Besides, the ebisu\_uit team maintains a stable level on the model they trained from the public test round to the private test round. They have kept 2nd place in the rankings with their F1 score of 77.22\%. Once again, we can see that the results are not much different between the 1st and 2nd place teams.

Furthermore, ebisu\_uit is also the team with the highest results on the EM measure with 67.43\%. If we take a look at the F-NLP team, it shows a similar trend with vc-tus. Remaining 5th in the public test round, their system helped them finish this task at 3rd with 76.46\% of F1 score. Generally, all the teams in this round were having trouble with the private test set since its difficulty had increased significantly. As a result, the submission results of the teams are reduced considerably compared to the public test round.

%% file: section/6-resultanalysis.tex
\section{Result Analysis}
\label{resultanalysis}
To gain a deeper insight into machine reading comprehension and question answering in Vietnamese, we analyze the results based on the 5 most powerful models at the VLSP 2021 - ViMRC Challenge.

\begin{figure*}[h]
    \centering
    
    \begin{minipage}{1.0\textwidth}
        \centering
        \includegraphics[width=0.7\linewidth, height=0.2\textheight]{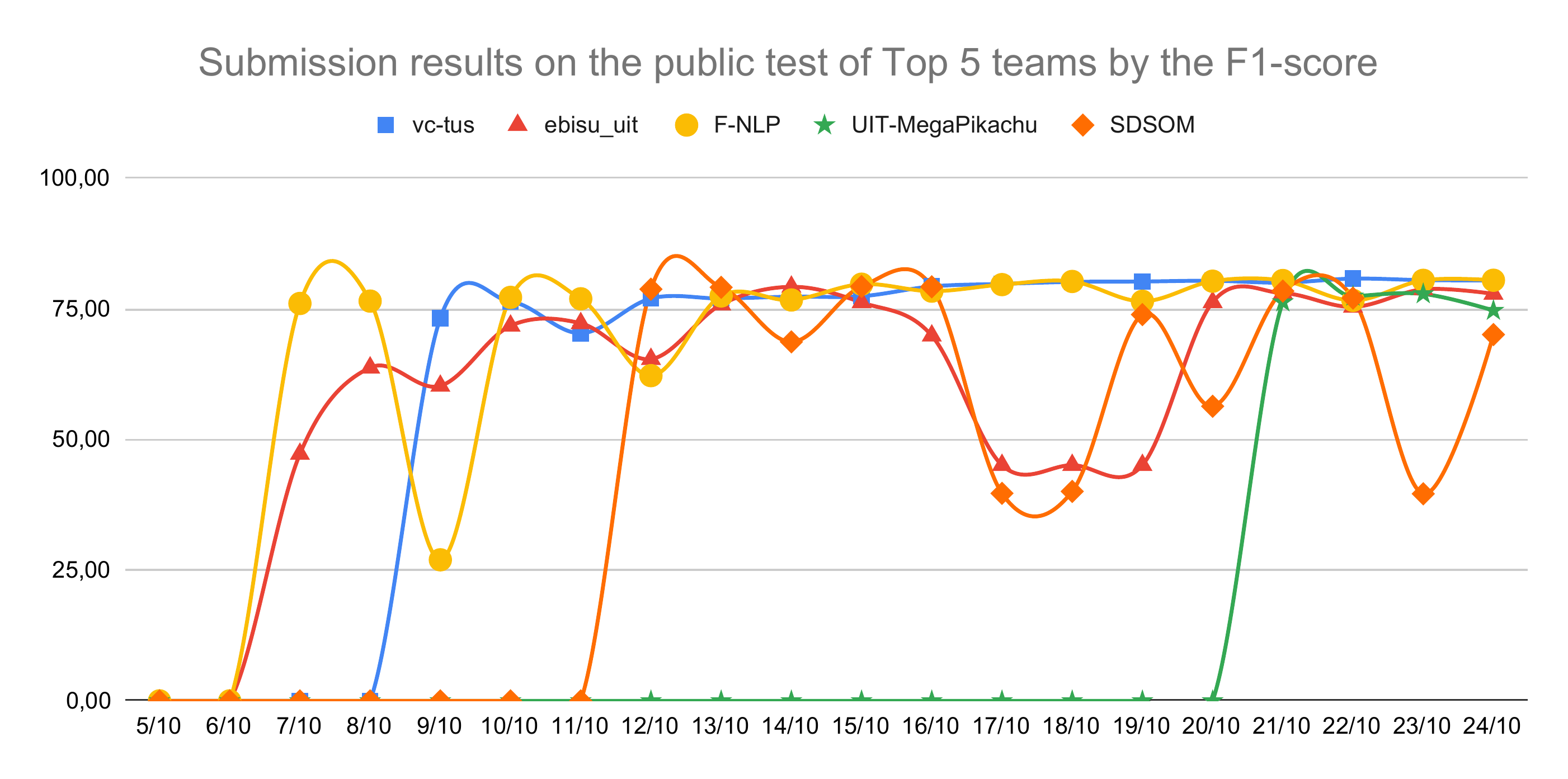}
    \end{minipage}
    \begin{minipage}{1.0\textwidth}
        \centering
        \includegraphics[width=0.7\linewidth, height=0.2\textheight]{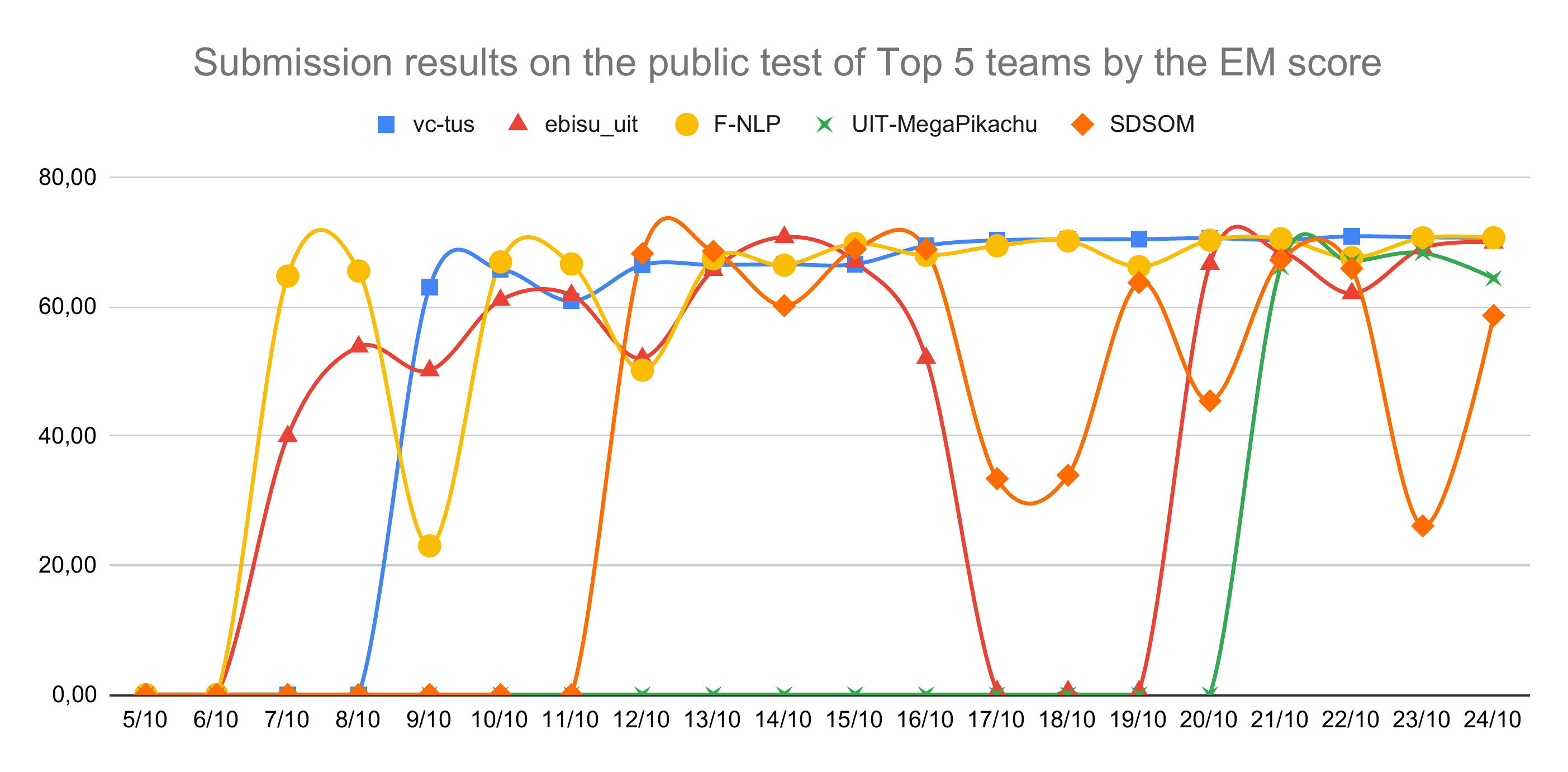}
    \end{minipage}
    \caption{Submission progress of the Top 5 teams on the public test phase.}
    \label{fig_submission_progress_public}
\end{figure*}

\begin{figure*}[ht]
    \centering
    
    \begin{minipage}{1.0\textwidth}
        \centering
        \includegraphics[width=0.83\linewidth, height=0.25\textheight]{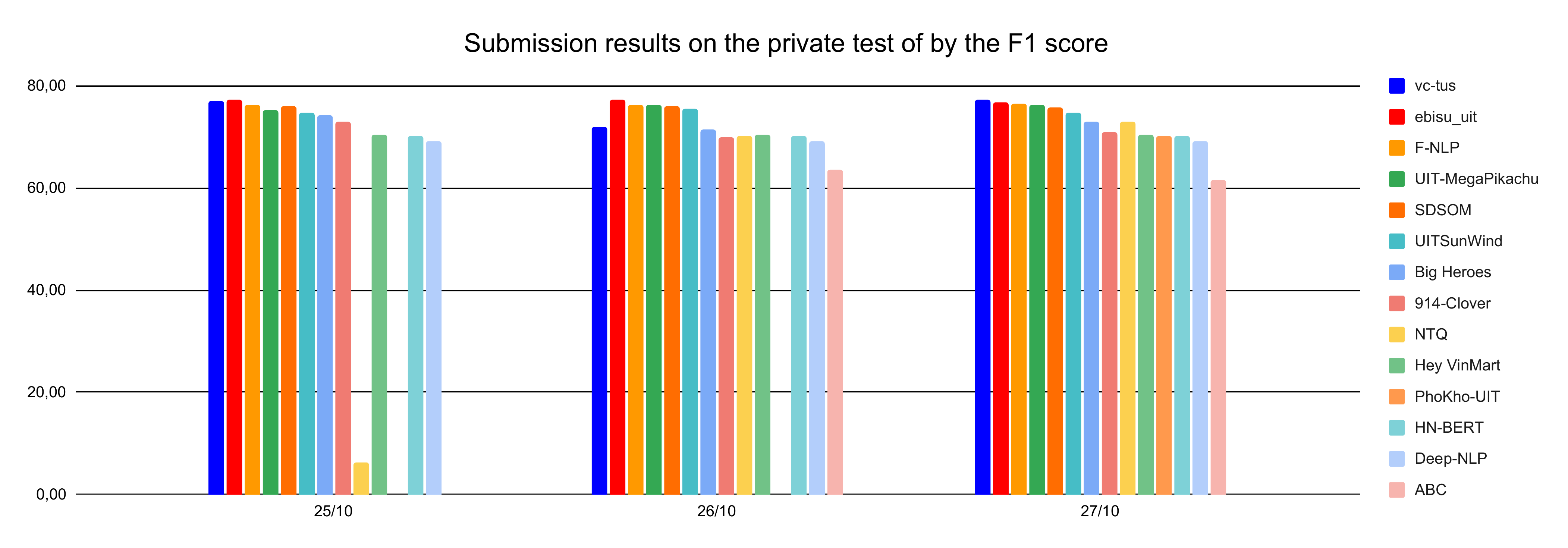}
    \end{minipage}
    \begin{minipage}{1.0\textwidth}
        \centering
        \includegraphics[width=0.83\linewidth, height=0.25\textheight]{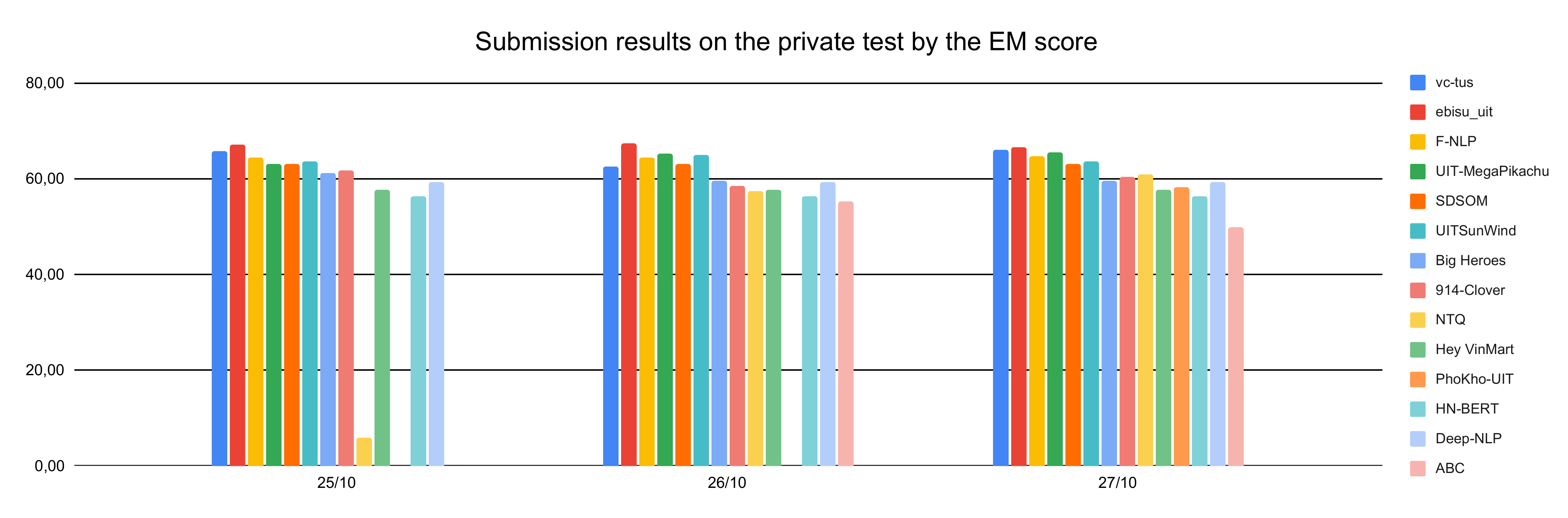}
    \end{minipage}
    \caption{Submission progress of the teams who results are higher than baseline score on the private test phase.}
    \label{fig_submission_progress_private}
\end{figure*}

\begin{table*}[h]
\centering

\resizebox{\textwidth}{!}{
\begin{tabular}{llrrrrrr}
\hline
\multicolumn{1}{c}{}                                         & \multicolumn{1}{c}{}                                  & \multicolumn{2}{c}{\textbf{Answerable}}                            & \multicolumn{2}{c}{\textbf{Unanswerable}}                          & \multicolumn{2}{c}{\textbf{Overall}}                               \\ \cline{3-8} 
\multicolumn{1}{c}{\multirow{-2}{*}{\textbf{Teams}}}         & \multicolumn{1}{c}{\multirow{-2}{*}{\textbf{Models}}} & \multicolumn{1}{c}{\textbf{EM}} & \multicolumn{1}{c}{\textbf{F1}} & \multicolumn{1}{c}{\textbf{EM}} & \multicolumn{1}{c}{\textbf{F1}} & \multicolumn{1}{c}{\textbf{EM}} & \multicolumn{1}{c}{\textbf{F1}} \\ \hline
vc-tus          & Retrospective Reader + XLM-R (Ensemble)                & \multicolumn{1}{r}{57.67}       & 73.54                            & \multicolumn{1}{r}{85.84}       & 85.84                            & \multicolumn{1}{r}{66.14}       & \bf \colorbox{green}{77.24}                            \\ 
ebisu\_uit      & BLANC + XLM-R/SemBERT (Ensemble)                       & \multicolumn{1}{r}{56.59}       & 70.59                            & \multicolumn{1}{r}{\bf \colorbox{green}{92.65}}       & \bf \colorbox{green}{92.65}                            & \multicolumn{1}{r}{\bf \colorbox{green}{67.43}}       & 77.22                            \\
F-NLP           & XLM-R (Ensemble)                                       & \multicolumn{1}{r}{58.78}       & \bf \colorbox{green}{75.66}                            & \multicolumn{1}{r}{78.32}       & 78.32                            & \multicolumn{1}{r}{64.66}       & 76.46                            \\ 
UIT-MegaPikachu & XLM-R (Single)                                         & \multicolumn{1}{r}{58.82}       & 74.63                            & \multicolumn{1}{r}{80.47}       & 80.47                            & \multicolumn{1}{r}{65.33}       & 76.39                            \\ 
UITSunWind      & XLM-R + BiLSTM (Ensemble)                                & \multicolumn{1}{r}{\bf \colorbox{green}{58.94}}       & 74.26                            & \multicolumn{1}{r}{78.67}       & 78.67                            & \multicolumn{1}{r}{64.87}       & 75.59                            \\ 
HN-BERT         & PhoBERT\_Large+R3F+CS (Single)                                  & \multicolumn{1}{r}{47.50}       & 66.99                            & \multicolumn{1}{r}{77.33}       & 77.33                            & \multicolumn{1}{r}{56.47}       & 70.10                            \\ \hline
Baseline                                & mBERT (Single)                                         & \multicolumn{1}{r}{41.72}       & 57.43                            & \multicolumn{1}{r}{67.11}       & 67.11                            & \multicolumn{1}{r}{49.35}       & 60.34                            \\ \hline
\end{tabular}
}
\caption{Final results on answerable and unanswerable questions of the private test set.}
\label{auaquestionsana}
\end{table*}

\subsection{Competition Progress Analysis}
Figure \ref{fig_submission_progress_public} illustrates the submission progress of the top 5 teams on the public test from October 5, 2021, to October 24, 2021. In this phase, we allow 10 submissions per day. However, according to Figure \ref{fig_submission_progress_public}, the submission results on both F1 and EM scores are not stable, which oscillates within the submission time. Besides, the results by EM score are no higher than 80\%, indicating the challenge in the dataset for the participants. 

In addition, Figure \ref{fig_submission_progress_private} illustrates the final submission results of participant teams. The private test started on October 25, 2021, and ended on October 27, 2021. Within 3 days of submission, the results on the F1 score do not change too much. Both F1 and EM scores achieved by participants are not higher than 80\% in this phase. Especially, for the final results, the team name \textbf{ebisu\_uit} has a lower result than the \textbf{vc\_tus} team but achieved a higher result on the EM score. It can be seen from the chart that the team \textbf{vc\_tus} achieved the best results by the F1-score, and the team \textbf{ebisu\_uit} achieved the best result by the EM score, which placed the $1^{st}$ and $2^{nd}$ in the competition.

\subsection{Answerable vs. Unanswerable Analysis}

To better understand the ability of the MRC systems to answer questions, we analyze human performance and the experimental results of the baseline model and the participating teams \cite{vlspmrc1,vlspmrc2,vlspmrc3,vlspmrc4,vlspmrc5,vlspmrc6}. Table \ref{auaquestionsana} shows final results on answerable and unanswerable questions of the private test set, evaluated on EM and F1 scores. As seen from the table, performances on unanswerable questions are always higher than on answerable questions. The ebisu\_uit team achieved the best performance on unanswerable questions with over 92\% of F1. However, the F-NLP and UIT-SunWind teams achieved the highest scores on the answers with 75.66\% of F1 and 58.82\% of EM, respectively. Interestingly, the vc-test team did not obtain the best performance on unanswerable and answerable questions, but this team achieved the best performance on the overall F1-score because they balanced the performances between the two types of questions better than the other teams.

\subsection{Challenging Question Examples}

\begin{figure*}
    \centering
    \includegraphics[width=1\linewidth]{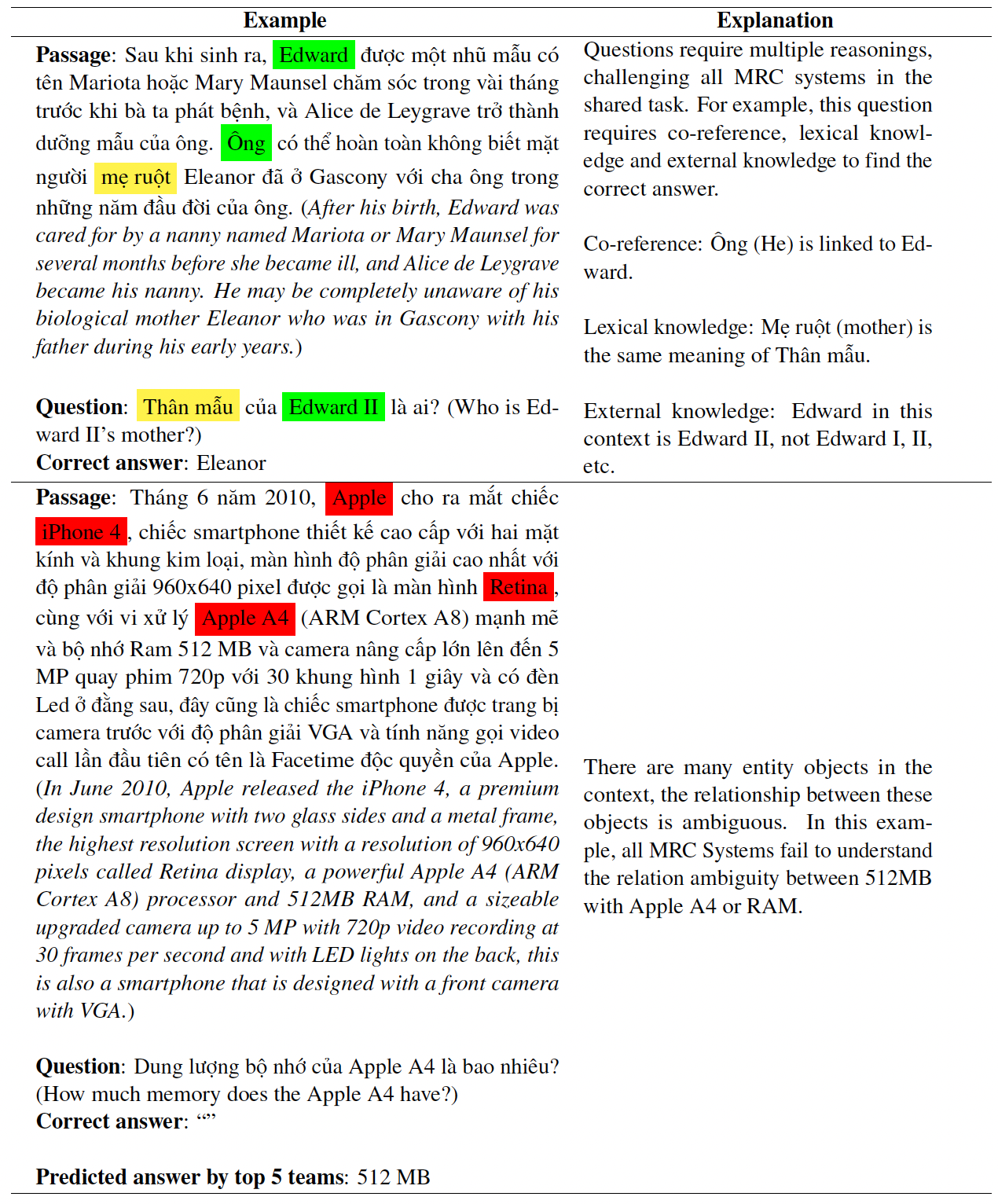}
    \caption{Several examples and explanations that the models failed to predict correct answers. The example texts in the UIT-ViQuAD 2.0 dataset are taken from the Vietnamese Wikipedia.}
    \label{failedwxample}
\end{figure*}

We select several typical examples of answerable and unanswerable questions that make it difficult for the models proposed by the teams \cite{vlspmrc1,vlspmrc2,vlspmrc3,vlspmrc4,vlspmrc5,vlspmrc6}. Figure \ref{failedwxample} presents several examples and explanations that the models failed to predict correct answers. We will explore more complex questions inspired by the works \cite{sugawara2018makes,sugawara2020assessing}.

%% file: section/7-conclusion.tex
\section{Conclusion and Future Work}
\label{conclusion}
The VLSP 2021 - ViMRC Challenge on Machine Reading Comprehension for Vietnamese has been organized at the VLSP 2021. Despite the fact that 77 teams had signed up to get the training datasets, only 24 teams were able to submit their results. Because several teams enrolled for many challenges at the VLSP 2021, the other teams may not have enough time to explore MRC models. This challenge provides valuable resources for developing Vietnamese machine reading comprehension, question answering, question generation (QG), and other AI applications using MRC, QA, and QG models.

To increase performance in machine reading comprehension systems, in the future, we intend to increase the amount and quality of annotated questions. In addition, we also make difficult questions based on findings proposed by the research works \cite{sugawara2017evaluation,sugawara2018makes,sugawara2021benchmarking}. UIT-ViQuAD 2.0 can also be used to evaluate various other NLP tasks: question answering that uses retriever-reader techniques \cite{chen2017reading,bertqas}, question generation \cite{du2017learning}, and information retrieval \cite{karpukhin2020dense}. We will explore more complex questions inspired by the works \cite{sugawara2018makes,sugawara2020assessing}. Finally, UIT-ViQuAD 2.0 will be provided to evaluate MRC, QA, and QG models, including the training set, the development set (public test set) and the test set (private test set).